\documentclass[sigconf, nonacm]{acmart}

\usepackage[T1]{fontenc}
\usepackage{soul}
\usepackage{url}
\usepackage{hyperref}
\usepackage[utf8]{inputenc}
\usepackage{caption}
\usepackage{graphicx}
\usepackage{tabularray}
\usepackage{amsmath}
\usepackage{booktabs}
\usepackage{algorithm}
\usepackage{algorithmic}
\usepackage{comment}
\usepackage{array}

\usepackage{xspace}
\usepackage{hhline}
\usepackage{makecell}
\usepackage{rotating}
\usepackage{multirow}
\usepackage{colortbl}
\usepackage{amsfonts}

\newcommand{\bfA}{\mathbf{A}}
\newcommand{\bfX}{\mathbf{X}}
\newcommand{\zuv}{\mathbf{Z}_{uv}}

\newcommand{\mr}{\mathbf{M}^{(i)}}
\newcommand{\mruv}{\mathbf{M}^{(i)}_{uv}}
\newcommand{\xruv}{\mathbf{X}^{(i)}_{uv}}
\newcommand{\zruv}{\mathbf{Z}^{(i)}_{uv}}

\newcommand{\posplus}{$\text{PoS}^{\text{+}}$\xspace}
\newcommand{\pos}{{P\MakeLowercase{o}S}\xspace}
\newcommand{\sop}{{S\MakeLowercase{o}P}\xspace}
\newcommand{\sopplus}{$\text{SoP}^{\text{+}}$\xspace}

\newcommand{\ssgrl}{\textit{S3GRL}\xspace}
\newcommand{\scaled}{S\MakeLowercase{ca}L\MakeLowercase{ed}\xspace}

\newcommand{\first}{\textcolor[rgb]{0.9, 0.17, 0.31}}
\newcommand{\second}{\textcolor[rgb]{0.0, 0.0, 1.0}}
\newcommand{\third}{\textcolor[rgb]{0.55, 0.0, 0.55}}
\newcommand{\worst}{\cellcolor[rgb]{0.957,0.8,0.8}}
\newcommand{\best}{\cellcolor[rgb]{0.851,0.918,0.827}}
\newcommand{\base}{\cellcolor[rgb]{1,0.949,0.8}}
\newcommand{\grA}[1]{#1\textsf{x}}

\DeclareMathOperator{\pool}{pool}

\AtBeginDocument{%
  \providecommand\BibTeX{{%
    \normalfont B\kern-0.5em{\scshape i\kern-0.25em b}\kern-0.8em\TeX}}}
    
\begin{document}

\title{Simplifying Subgraph Representation Learning \\ for Scalable Link Prediction}

\author{Paul Louis}
\email{paul.louis@ontariotechu.net}
\affiliation{%
  \institution{Ontario Tech University}
  \city{Oshawa}
  \state{Ontario}
  \country{Canada}
}

\author{Shweta Ann Jacob}
\email{shweta.jacob@ontariotechu.net}
\affiliation{%
  \institution{Ontario Tech University}
  \city{Oshawa}
  \state{Ontario}
  \country{Canada}
}

\author{Amirali Salehi-Abari}
\email{abari@ontariotechu.ca}
\affiliation{%
  \institution{Ontario Tech University}
  \city{Oshawa}
  \state{Ontario}
  \country{Canada}
}

\renewcommand{\shortauthors}{Louis et al.}

\begin{abstract}
Link prediction on graphs is a fundamental problem, addressing the task of predicting potential connections between entities in a network, enabling the identification of missing or future relationships. Its applications span various domains, including social networks, recommendation systems, and biological networks, where predicting links can enhance our understanding of network structure and facilitate more effective information dissemination or resource allocation. Subgraph representation learning approaches (SGRLs), by transforming link prediction to graph classification on the subgraphs around the links, have achieved state-of-the-art performance in link prediction. However, SGRLs are computationally expensive and not scalable to large-scale graphs due to expensive subgraph-level operations. To unlock the scalability of SGRLs, we propose a new class of SGRLs, that we call \textit{Scalable Simplified SGRL} (\ssgrl). Aimed at faster training and inference, \ssgrl simplifies the message passing and aggregation operations in each link's subgraph. \ssgrl, as a scalability framework, accommodates various subgraph sampling strategies and diffusion operators to emulate computationally expensive SGRLs. We propose multiple instances of \ssgrl and empirically study them on small to large-scale graphs. Our extensive experiments demonstrate that the proposed \ssgrl models scale up SGRLs without significant performance compromise (even with considerable gains in some cases), while offering substantially lower computational footprints (e.g., multi-fold inference and training speedup).
\end{abstract}


\settopmatter{printfolios=true}
\maketitle

\section{Introduction}

Graphs are ubiquitous in representing relationships between interacting entities in a variety of contexts, ranging from social networks \cite{chen2020friend} to polypharmacy \cite{zitnik2018modeling}. One of the main tasks on graphs is link prediction \cite{liben2003link}, which involves predicting future or missing relationships between pairs of entities, given the graph structure. Link prediction plays a fundamental role in impactful areas, including molecular interaction predictions \cite{huang2020skipgnn}, recommender systems \cite{ying2018graph}, and protein-protein interaction predictions \cite{zhong2022long}.

The early solutions to link prediction relied on hand-crafted, parameter-free heuristics based on the proximity of the source-target nodes \cite{page1999pagerank,adamic2003friends}.
However, these methods require extensive domain knowledge for effective implementation. Recently, the success of \textit{graph neural networks (GNNs)} in learning latent
representations of nodes \cite{kipf2017semi} has led to their application in link prediction, through aggregating the source-target nodes' representations as a link representation \cite{kipf2016variational,pan2018adversarially}. However, as GNNs learn a pair of nodes' representations independent of their relative positions to each other, aggregating independent node representations results in poor link representations \cite{zhang2021labeling}. To circumvent this, the state-of-the-art subgraph representation learning approaches (SGRLs) \cite{zhang2018link} learn enclosing subgraphs of pairs of source-target nodes, while augmenting node features with structural features. By doing so, the link prediction problem is cast as a binary graph classification on the enclosing subgraph about a link. 

The core issue hampering the deployment of GNNs and SGRLs is their high computational demands on large-scale graphs (e.g., OGB datasets \cite{hu2020ogb,hu2021ogb}). Many approaches have been proposed for GNNs to relieve this bottleneck, ranging from sampling \cite{chen2018fastgcn,graphsaint-iclr20} to simplification \cite{wu2019simplifying} techniques. However, these techniques cannot be directly applied in SGRL models, where the issue is even more adverse due to subgraph extractions pertaining to each link. Although recent work has focused on tackling SGRL scalability by sampling \cite{yin2022algorithm,louis2022sampling} or sketching \cite{chamberlain2023graph} approaches, there is a lack of simplification techniques specifically designed for SGRLs. We focus on improving the scalability in SGRLs by simplifying the underlying training mechanism.

The motivation for our work stems from the increasing need for scalable and efficient link prediction methods in large-scale graphs. Traditional link prediction approaches often struggle with scalability and computational efficiency when applied to these massive datasets. Our Simplified subgraph representation learning approach offers a promising solution by reducing computational complexity while maintaining predictive performance.

Inspired by the simplification techniques in GNNs such as SGCN \cite{wu2019simplifying} and SIGN \cite{sign_icml_grl2020}, we propose \textit{Scalable Simplified  SGRL} (\ssgrl) framework, which extends the simplification techniques to SGRLs. Our \ssgrl is flexible in emulating many SGRLs by offering choices of subgraph sampling and diffusion operators, producing different-sized convolutional filters for each link's subgraph. Our \ssgrl benefits from precomputing the subgraph diffusion operators, leading to an overall reduction in runtimes while offering a multi-fold scale-up over existing SGRLs in training/inference times. We propose and empirically study multiple instances of our \ssgrl framework. Our extensive experiments show substantial speedups in \ssgrl models over state-of-the-art SGRLs while maintaining and sometimes surpassing their efficacies. The key contributions are summarized as follows:
\begin{itemize}
    \item We introduce the \ssgrl framework,  extending simplification methods from GNNs to the context of subgraph representation learning, addressing the scalability issue in SGRLs for link prediction.
    \item Focusing on the link prediction problem, we introduce multiple instantiations of our \ssgrl to showcase the flexibility, robustness, and scalability of our framework. We theoretically analyze the inference time complexity, preprocessing time complexity, and storage requirement of our methods and contrasted them with their competitive counterparts. 
    \item We empirically study and evaluate the performance of our \ssgrl instances on 14 small-to-large graph datasets under various evaluation criteria. Our empirical studies and analyses include deep comparisons of our methods against 16 competitive baselines widely utilized in link prediction studies. Our empirical results confirm the efficacy improvements of our \ssgrl methods over others while offering multi-fold speed-up in training and inference. Our results also demonstrate a multi-fold reduction in storage requirements: The \ssgrl preprocessed datasets are multiple magnitudes smaller than those of other SGRLs (e.g., SEAL \cite{zhang2018link}).
    \item We contribute to the research community by sharing our findings and framework as an open-source framework\footnote{\href{https://github.com/venomouscyanide/S3GRL\_OGB}{https://github.com/venomouscyanide/S3GRL\_OGB}}. This enables other researchers to devise and test their own \ssgrl instances, fostering further advancements in link prediction research.
\end{itemize}

\section{Related Work}
\label{sec:relwork}
Graph representation learning (GRL) \cite{hamilton2020graph} has numerous applications in drug discovery \cite{xiong2019pushing}, knowledge graph completion \cite{zhang2020relational}, and recommender systems \cite{ying2018graph}. The key downstream tasks in GRL are node classification \cite{hamilton2017inductive}, link prediction \cite{zhang2018link} and graph classification \cite{zhang2018end}. The early work in GRL, \textit{shallow encoders} \cite{perozzi2014deepwalk,grover2016node2vec}, learns dense latent node representations by taking multiple random walks rooted at each node. However, shallow encoders were incapable of consuming node features and being applied in the inductive settings. This shortcoming led to growing interest in Message Passing Graph Neural Networks (MPGNNs)\footnote{We use MPGNNs and GNNs interchangeably even though GNNs represent the broader class in GRL.}
\cite{brunaspectral2014,defferrard2016convolutional} such as Graph Convolutional Networks (GCN) \cite{kipf2017semi}. In MPGNNs, node feature representations are iteratively updated by first aggregating neighborhood features and then updating them through non-linear transformations. MPGNNs differ in formulations of aggregation and update functions \cite{hamilton2020graph}.

Link prediction is a fundamental problem on graphs, where the objective is to compute the likelihood of the presence of a link between a pair of nodes (e.g., users in a social network) \cite{liben2003link}. Network heuristics such as Common Neighbors or Katz Index \cite{katz1953new} were the first solutions for link prediction. However, these predefined heuristics fail to generalize well on different graphs. To address this, MPGNNs have been used to learn representations of node pairs independently, then aggregate the representations for link probability prediction \cite{kipf2016variational,Mavromatis2021GraphIM}. However, this class of MPGNN solutions falls short in capturing graph automorphism and different structural roles of nodes in forming links \cite{zhang2021labeling}. SEAL \cite{zhang2018link} successfully overcomes this limitation by casting the link prediction problem as a binary graph classification on the enclosing subgraph about a link, and adding structural labels to node features. This led to the emergence of subgraph representation learning approaches (SGRLs) \cite{zhang2018link,li2020distance}, which offer state-of-the-art results on the link prediction task. However, a common theme impeding the practicality and deployment of SGRLs is its lack of scalability to large-scale graphs.

Recently, a new research direction has emerged in GNNs on scalable training and inference techniques, which mainly focuses on node classification tasks. A common practice is to use different sampling techniques---such as node sampling \cite{hamilton2017inductive}, layer-wise sampling \cite{zou2019layer}, and graph sampling \cite{graphsaint-iclr20}---to subsample the training data to reduce computation. Other approaches utilize simplification techniques (e.g., removing intermediate non-linearities \cite{wu2019simplifying,sign_icml_grl2020,saber2024}), hashing techniques \cite{10.1145/3442381.3449884}, or approximate search techniques \cite{pmlr-v198-wang22a} to speed up the training and inference. These approaches have been shown to fasten learning of MPGNNs (at the global level) for node and graph classification tasks, and even for link prediction \cite{10.1145/3546157.3546163,pmlr-v198-wang22a,10.1145/3442381.3449884}; however, they are not directly applicable to SGRLs which exhibit superior performance for link prediction.  
In this work, we take the scalability-by-simplification approach for SGRLs (rather than for MPGNNs) by introducing \textit{Scalable Simplified  SGRL} (\ssgrl). Allowing diverse definitions of subgraphs and subgraph-level diffusion operators for SGRLs, \ssgrl emulates SGRLs in generalization while speeding them up.

There is a growing interest in the scalability of SGRLs \cite{yin2022algorithm,louis2022sampling,chamberlain2023graph}, with a main emphasis on the sampling approaches.  SUREL \cite{yin2022algorithm}, by deploying random walks with restart, approximates subgraph structures; however, it does not use GNNs for link prediction. \scaled \cite{louis2022sampling} uses similar sampling techniques of SUREL, but to sparsify subgraphs in SGRLs for better scalability. ELPH/BUDDY \cite{chamberlain2023graph} uses subgraph sketches as messages in MPGNNs but does not learn link representations explicitly at a subgraph level. Most recently, sampling techniques have improved scalability in subgraph classification tasks \cite{jacob2023stochastic}. 
Our work complements this body of research. Our framework can be combined with (or host) these methods for better scalability (see our experiments for an example).

\section{Preliminaries}

Consider a graph $G = (V, E)$, where $V=\{1, \dots,n\}$ is the set of nodes (e.g., individuals, proteins), $E \subseteq V\times V$ represents their relationships (e.g., friendship or protein-to-protein interactions). We also let $\bfA \in \mathbb{R}^{n\times n}$
represent the \emph{adjacency matrix} of $G$, where $A_{uv} \neq 0$ if and only if  $(u,v) \in E$. 
We further assume each node possesses a $d$-dimensional feature vector (e.g., user's profile, features of proteins, etc.) stored as a row in \emph{feature matrix} $\bfX \in \mathbb{R}^{n\times d}$.  

\vskip 1.5mm
\noindent \textbf{Link Prediction.} The goal 
is to infer the presence or absence of an edge between \emph{target nodes} $T=\{u,v\}$ given the 
observed matrices $\bfA$ and $\bfX$. The learning problem is to find a \emph{likelihood function} $f$ such that it assigns a higher likelihood to those target nodes with a missing link. The likelihood function can be formulated by various neural network architectures (e.g., MLP \cite{guo2022linkless} or GNNs \cite{davidson2018hyperspherical}).

\vskip 1.5mm
\noindent \textbf{GNNs.} Graph Neural Networks (GNNs) typically take as input $\bfA$ and $\bfX$, apply $L$ layers of convolution-like operations, and output $\mathbf{Z} \in \mathbb{R}^{n\times d'}$, containing $d'$-dimensional representation for each node as a row. For example, the $l^{th}$-layer output of Graph Convolution Network (GCN) \cite{kipf2017semi} is given by 

\begin{equation}
    \mathbf{H}^{(l)} = \sigma \left(\hat{\bfA}\mathbf{H}^{(l - 1)}\mathbf{W}^{(l - 1)}\right),
    \label{eq:gcn}
\end{equation}
where \emph{normalized adjacency matrix} $\hat{\bfA}=\tilde{\mathbf{D}}^{-\frac{1}{2}}\tilde{\bfA}\tilde{\mathbf{D}}^{-\frac{1}{2}}$, $\tilde{\bfA} = \bfA + \mathbf{I}$, $\mathbf{I}$ is an identity matrix, $\tilde{\mathbf{D}}$ is the diagonal matrix with $\tilde{D}_{ii} = \sum_j \tilde{A}_{ij}$, and $\sigma$ is a non-linearity function (e.g., ReLU). Here, $\mathbf{H}^{(0)} =  \bfX$ and $\mathbf{W}^{(l)}$ is the $l^{th}$ layer's learnable weight matrix. After $L$ stacked layers of Eq.~\ref{eq:gcn}, GCN outputs  nodal-representation matrix $\mathbf{Z}= \mathbf{H}^{(L)}$. For any target nodes $T=\{u,v\}$, one can compute its joint representation:
\begin{equation}
    \mathbf{q}_{uv} = \pool\left(\mathbf{z}_{u}, \mathbf{z}_{v}\right),
    \label{eq:quv}
\end{equation}
where $\mathbf{z}_{u}$, $\mathbf{z}_{v}$ are $u$'s and $v$'s learned representations in $\mathbf{Z}$, and $\pool$ is a pooling/readout operation (e.g., Hadamard product or concatenation) to aggregate the pairs' representations. Then, the link probability $P_{uv}$ for the target is given by  $P_{uv} = \Omega(\mathbf{q}_{uv})$, where $\Omega$ is a learnable non-linear function (e.g., MLP) transforming the target embedding $\mathbf{q}_{uv}$ into a link probability.

\begin{figure*}[t]
  \centering
  \includegraphics[width=\textwidth]{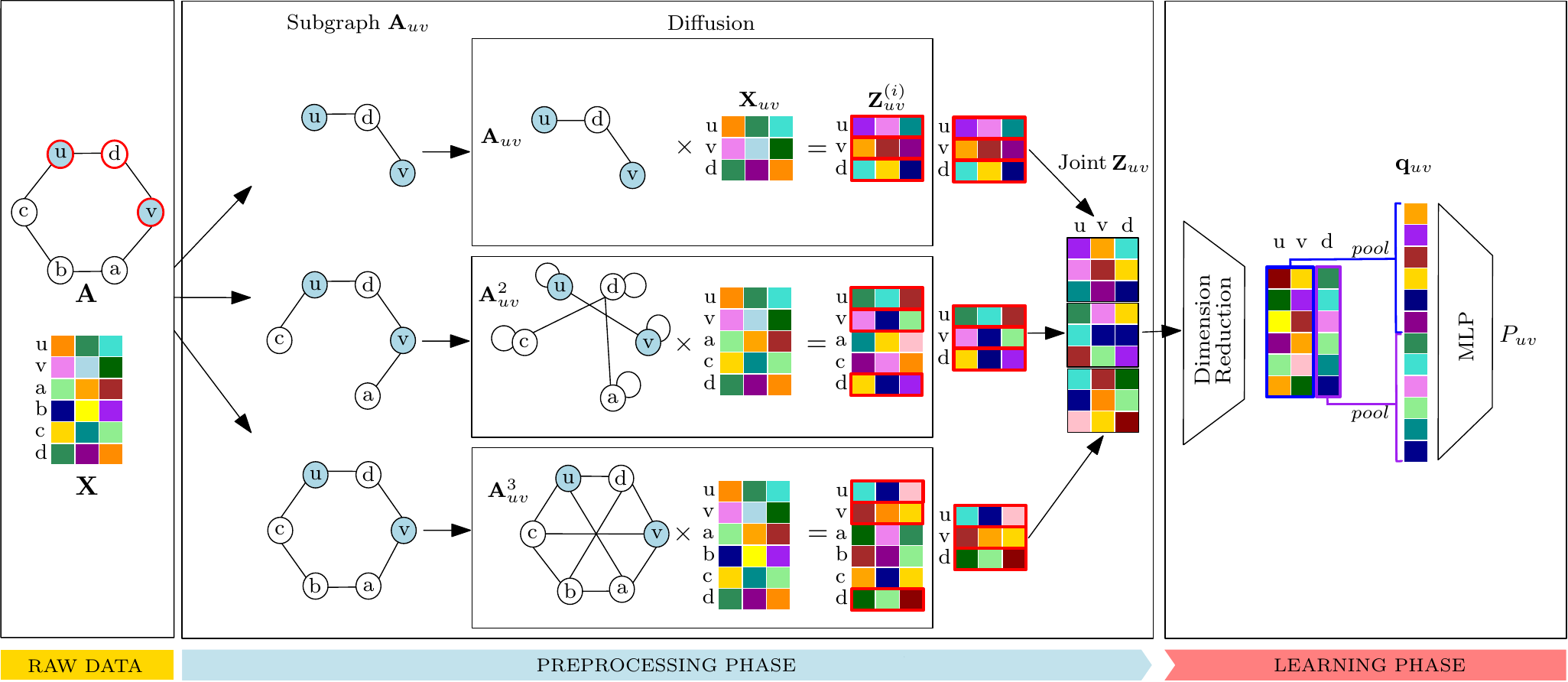}
  \caption{Our \ssgrl framework: In the preprocessing phase (shown by the shaded blue arrow), first multiple subgraphs are extracted around the target nodes $u$ and $v$ (shaded in blue) by various sampling strategies.  Diffusion matrices are then created from extracted subgraph adjacency matrices by predefined diffusion operators (e.g., powers of subgraphs in this figure). Each diffusion process involves the application of the subgraph diffusion matrix on its nodal features to create the matrix $\zruv$. The operator-level node representations of selected nodes (with a red border in raw data) are then aggregated for all subgraphs to form the joint $\zuv$ matrix. The selected nodes in this example are the target nodes $\{u,v\}$, and their common neighbor $d$. In the learning phase (as shown by the shaded red arrow), the joint matrix $\zuv$ undergoes dimensionality reduction followed by pooling using center pooling (highlighted by blue-border box) and common neighbor pooling (highlighted by purple-border box). Finally, the target representation $\textbf{q}_{uv}$ is transformed by an MLP to a link probability $P_{uv}$.}
 
  \label{fig:architecture}
\end{figure*}

\vskip 1.5mm
\noindent \textbf{SGRLs.} Subgraph representation learning approaches (SGRLs)  \cite{zhang2017weisfeiler,zhang2018link,zhang2021labeling,yin2022algorithm,louis2022sampling} treat link prediction as a binary graph classification problem on enclosing subgraph $G_{uv} = (V_{uv} \subseteq V, E_{uv} \subseteq E)$ around a target $\{u,v\}$. SGRLs aim to classify if the enclosing subgraph $G_{uv}$ is \textit{closed} or \textit{open} (i.e., the link exists or not). For each $G_{uv}$, SGRLs produce its nodal-representation matrix $\mathbf{Z}_{uv}$ using the stack of graph convolution operators (see Eq.~\ref{eq:gcn}). Then, similar to Eq.~\ref{eq:quv}, the nodal representations would be converted to link probabilities, with the distinction that the pooling function is graph pooling (e.g., SortPooling \cite{zhang2018end}), operating over all node's representations (not just those of targets) to produce the fixed-size representation of the subgraph. To improve the expressiveness of GNNs on subgraphs, SGRLs augment node features with some structural features determined by the targets' positions in the subgraph. These augmented features are known as \textit{node labels} \cite{zhang2021labeling}. Node labeling techniques fall into \emph{zero-one} or \textit{geodesic distance-based} schemes \cite{huang2022boosting}. We refer to $\bfX_{uv}$ as the matrix containing both the initial node features and labels generated by a valid labeling scheme \cite{zhang2021labeling}. The expressiveness power of SGRLs comes with high computational costs due to subgraph extractions, node labeling, and independent operations on overlapping large subgraphs. This computational overhead is exaggerated in denser graphs and deeper subgraphs due to the exponential growth of subgraph size. 

\vskip 1.5mm
\noindent \textbf{SGCN and SIGN.} To improve the scalability of GNNs for node classification tasks, several attempts are made to simplify GNNs by removing their intermediate nonlinearities, thus making them shallower, easier to train, and scalable. Simplified GCN (SGCN) \cite{wu2019simplifying} has removed all intermediate non-linearities in L-layer GCNs, to predict class label probabilities $\mathbf{Y}$ for all nodes by
$\mathbf{Y}= \xi (\hat{\bfA}^{L}\mathbf{X}\mathbf{W})$,
where $\xi$ is softmax or logistic function, and $\mathbf{W}$ is the only learnable weight matrix. The term $\hat{\bfA}^{L}\mathbf{X}$ can be precomputed once before training. Benefiting from this precomputation and a shallower architecture, SGCN improves scalability. SIGN has extended SGCN to be  more expressive yet scalable for node classification tasks. Its crux is to deploy a set of linear diffusion matrices $\mathbf{M}^{(1)},\dots,\mathbf{M}^{(r)}$ that can be applied to node-feature matrix $\bfX$. The class probabilities $\mathbf{Y}$ are computed by  $\mathbf{Y} = \xi (\mathbf{Z}\mathbf{W}')$, where  
\begin{equation}
\mathbf{Z} = \sigma \left(\bigoplus_{i=0}^{r}\mr\bfX\mathbf{W}_{i} \right).
\label{eq:sign_op_learning}
\end{equation}
Here, $\xi$ and $\sigma$ are non-linearities, $\mathbf{W}_{i}$ is the learnable weight matrix for diffusion operator $\mr$, $\bigoplus$ is a concatenation operation, and $\mathbf{W}'$ is the learnable weight matrix for transforming node representations to class probabilities. Letting $\mathbf{I}$ be the identity matrix,  $\mathbf{M}^{(0)} = \mathbf{I}$ in Eq.~\ref{eq:sign_op_learning} allows the node features to contribute directly (independent of graph structure) into the node representation and consequently to the class probabilities. The $\mr$ operator captures different heuristics in the graph. 
As with SGCN, the terms $\mr\bfX$ in SIGN can be once precomputed before training. These precomputations and the shallow architecture of SIGN lead to substantial speedup during training and inference with limited compromise on node classification efficacy. Motivated by these efficiencies, our \ssgrl framework extends SIGN for link prediction on subgraphs while addressing the computational bottleneck of SGRLs.

\section{Scalable Simplified  SGRL (\ssgrl)}
We propose \textit{Scalable Simplified  SGRL} (\ssgrl), which benefits from the expressiveness power of SGRLs while offering the simplicity and scalability of SIGN and SGCN for link prediction. Our framework leads to a multi-fold speedup in both training and inference of SGRL methods while maintaining or boosting their state-of-the-art performance (see experiments below). 

Our \ssgrl framework consists of two key components: (i) \emph{Subgraph sampling strategy} $\Psi(G,T)$ takes as an input the graph $G$ and target pairs $T=\{u,v\}$ and outputs the adjacency matrix $\mathbf{A}_{uv}$ of the enclosing subgraphs $G_{uv}$ around the targets. The subgraph sampling strategy $\Psi$ can capture various subgraph definitions such as $h$-hop enclosing subgraphs \cite{zhang2018link}), random-walk-sampled subgraphs \cite{yin2022algorithm,louis2022sampling}, and heuristic-based subgraphs \cite{zeng2021decoupling}; (ii) \emph{Diffusion operator} $\Phi(\mathbf{A}_{uv})$  takes the subgraph adjacency matrix $\mathbf{A}_{uv}$ and outputs its diffusion matrix $\mathbf{M}_{uv}$. A different class of diffusion operators are available: adjacency/Laplacian operators to capture connectivity, triangle/motif-based operators \cite{granovetter1983strength} to capture inherent community structure, personalized pagerank-based (PPR) operators \cite{gasteiger2019diffusion} to identify important connections. Each of these operators and their powers can constitute the different diffusion operators in \ssgrl.

In our \ssgrl framework, each model is characterized by the \emph{sampling-operator} set $\mathcal{S}=\{(\Psi_i, \Phi_i)\}_{i=0}^r$, where $\Psi_i$ and $\Phi_i$  are the $i^{th}$ subgraph sampling strategy and diffusion operator, respectively. For a graph $G$, target pair $T=\{u,v\}$, and  sampling-operator pair $(\Psi_i, \Phi_i)$, one can find $T$'s sampled subgraph $\bfA^{(i)}_{uv}=\Psi_i(G,T)$ and its corresponding diffusion matrix $\mruv=\Phi_i(\bfA^{(i)}_{uv})$. For instance, one can define $\Psi_{i}$ to sample the random-walk-induced subgraph $\mathbf{A}_{uv}^{(i)}$ rooted at $\{u,v\}$ in $G$. Then, $\Phi_{i}$ can compute the $l$-th power of $\mathbf{A}_{uv}^{(i)}$, to count the number of $l$-length walks on the subgraph. \ssgrl computes the operator-level node representations of the selected subgraph $\bfA^{(i)}_{uv}$ by 
\begin{equation}
\zruv = \mruv \xruv.
\label{eq:operator-rep}
\end{equation} 
Here, $\xruv$ is the node feature matrix for nodes in the selected subgraph $\bfA^{(i)}_{uv}$ for the sampling-operator pair $(\Psi_i, \Phi_i)$. Eq.~\ref{eq:operator-rep} can be viewed as feature smoothing where the diffusion matrix $\mruv$ is applied over node features $\xruv$. \ssgrl then concatenates the operator-level nodal-representation matrix $\zruv$ of all sampling-operator pairs to form the \textit{joint nodal-representation matrix}:
\begin{equation}
\mathbf{Z}_{uv}  = \bigoplus_{i=0}^r \zruv
\label{eq:joint-rep}    
\end{equation}
The concatenation between nodal-representation matrices with dimensionality mismatch should be done with care: the corresponding rows (belonging to the same node) should be inline, where missing rows are filled with zeros (analogous to zero-padding for graph pooling). The joint nodal-representation matrix $\mathbf{Z}_{uv}$ goes through a non-linear feature transformation (for dimensionality reduction) by learnable weight matrix $\mathbf{W}$ and non-linearity $\sigma$. This transformed matrix is then further downsampled by the graph pooling $\pool$ to form the target's representation:
\begin{equation}
\mathbf{q}_{uv} = \pool\left(\sigma\left(\mathbf{Z}_{uv} \mathbf{W}\right)\right),
\label{eq:simple_sgrl}
\end{equation} 
from which the link probability is computed by 
 \begin{equation}
 P_{uv} = \Omega(\mathbf{q}_{uv}),    
\label{eq:link-prob-s3grl}
 \end{equation}
with $\Omega$ being a learnable non-linear function (e.g., MLP) to convert the target representation $\mathbf{q}_{uv}$ into a link probability.

Our \ssgrl exhibits substantial speedup in inference and training through precomputing $\mathbf{Z}_{uv}$ in comparison to more computationally-intensive SGRLs  \cite{li2020distance,pan2022neural}, designed based on multi-layered GCN or DGCNN \cite{zhang2018end} (see our experiments for details). Note that only Equations \ref{eq:simple_sgrl} and \ref{eq:link-prob-s3grl} are utilized in training and inference in \ssgrl (see also Figure \ref{fig:architecture}).  Apart from this computational speedup, \ssgrl offers other advantages analogous to other prominent GNNs:

\vskip 1.5mm
\noindent \textbf{Disentanglement of Data and Model.} The composition of subgraph sampling strategy $\Psi$ and diffusion operator $\Phi$ facilitates the disentanglement (or decoupling) of data (i.e., subgraph) and model (i.e., diffusion operator). This disentanglement resembles shaDow-GNN \cite{zeng2021decoupling}, in which the depth of GNNs (i.e., its number of layers) is decoupled from the depth of enclosing subgraphs. Similarly, in \ssgrl, one can explore high-order diffusion operators (e.g., adjacency matrix to a high power) in constrained enclosing subgraphs (e.g., ego network of the target pair). This combination simulates multiple message-passing updates between the nodes in the local subgraph, thus achieving local oversmoothing  \cite{zeng2021decoupling} and ensuring the final subgraph representation possesses information from all the nodes in the local subgraph.

\vskip 1.5mm
\noindent \textbf{Multi-View Representation.} Our \ssgrl framework via the sampling-operator pairs provides multiple views of the enclosing neighborhood of each target pair. This capability allows hosting models analogous to multi-view \cite{abu2020n,Cai_Ji_2020} models.

\section{\ssgrl Instances and Computational Complexities}
We introduce two instances of \ssgrl, differentiating in the choice of 
sampling-operator pairs, which we call \textit{Powers of Subgraphs (\pos)} and  \textit{Subgraphs of Powers (\sop).} We then present our graph pooling methods for these two instances. Finally, we deep dive into computational time complexity analyses of these variants and compare them with their competitive counterparts SEAL \cite{zhang2018link}.

\subsection{Powers of Subgraphs (\pos)}
This instance intends to mimic a group of SGRLs with various model depths on a fixed-sized enclosing subgraph while boosting scalability and generalization (e.g., SEAL \cite{zhang2018link}). The sampling-operator set $\mathcal{S} = \{(\Psi_{i}, \Phi_{i})\}_{i=0}^r$ is defined as follows: (i) the sampling strategy $\Psi_{i}(G, T)$ for any $i$ is constant and returns the adjacency matrix $\bfA_{uv}$ of the $h$-hop enclosing subgraph $G_{uv}^h$ about target $T=\{u,v\}$. The subgraph $G_{uv}^h$ is a node-induced subgraph of $G$ with the node set $V^h_{uv} = \{j | d(j,u) \leq h\ or\ d(j,v) \leq h \}$, including the nodes with the geodesic distance of at most $h$ from either of the target pairs \cite{zhang2018link}; (ii) the $i$-th diffusion operator $\Phi_{i}(\bfA) = \mathbf{A}^{i}$ is the $i$-th power of adjacency matrix $\bfA$. This operator facilitates information diffusion among nodes $i$-length paths apart. Putting (i) and (ii) together, one can derive $\mruv = \bfA_{uv}^i$ for Eq.~\ref{eq:operator-rep}.  Note that $\mathbf{M}^{(0)}_{uv} =$ \textbf{I} allows the node features to contribute directly (independent of graph structure) to subgraph representation. \pos possesses two hyperparameters $r$ and $h$, controlling the number of diffusion operators and the number of hops in enclosing subgraphs, respectively. 

One can intuitively view \pos as equivalent to SEAL that uses a GNN model of depth $r$ with ``skip-connections" \cite{xu2018representation} on the $h$-hop enclosing subgraphs. The skip-connections property allows consuming all the intermediate learned representations of each layer to construct the final link representation. Similarly, \pos uses varying $i$-th power diffusion operators combined with the concatenation operator in Eq.~\ref{eq:simple_sgrl} to generate the representation $\mathbf{q}_{uv}$. In this light, the two hyperparameters of $r$ and $h$ (resp.) in \pos control the (virtual) depth of the model and the number of hops in enclosing subgraphs (resp.).

\begin{table*}[tb]
\centering
\caption{Computational time complexity: $n$ and $m$ are the number of
nodes and edges (resp.); $d$ and $d'$ are the dimensionality of node features and hidden layers (resp.); $L$ is the number of hidden layers for graph convolutions; and $k$ is the number of $l$-lengh random walks.}
\label{tab:time_complexity}
\vspace{-8pt}
\begin{tabular}{lcccc}
\toprule
 & SEAL Dyn. & SEAL Static & POS & POS+ScaLed \\
\midrule
Preprocessing & $O(1)$ & $O(m)$& $O(rnm + rmd)$& $O(rk^2l^2+rkld)$\\
Inference & $O(Lmdd'+d'^2)$ &  $O(Lmdd'+d'^2)$& $O(rdd'+d'^2)$ & $O(rdd'+d'^2)$\\
\bottomrule
\end{tabular}
\vspace{-6pt}
\end{table*}
\label{sec:pos}

\subsection{Subgraphs of Powers (\sop)}
This instance of \ssgrl, by transforming the global input graph G, brings long-range interactions into the local enclosing subgraphs. Let $\Psi_H(G, T, h)$ be \textit{h-hop sampling strategy}, returning the enclosing $h$-hop subgraph about the target pair $T=\{u,v\}$. The \sop model defines $\mathcal{S} = \{(\Psi_{i}, \Phi_{i})\}_{i=0}^r$ with $\Psi_i(G, T) = \Psi_H(G^i, T, h)$ and $\Phi_{i}(\bfA_{uv})=\bfA_{uv}$. Here, $G^i$ is the $i$-th power of the input graph (computed by $\bfA^i$), in which two nodes are adjacent when their geodesic distance in G is at most $i$. In this sense, the power of graphs brings long-range interactions to local neighborhoods at the cost of neighborhood densification. However, as the diffusion operator is an identity function, \sop prevents overarching to the further-away information of indirect neighbors. \sop consists of two hyperparameters $r$ and $h$ that control the number of diffusion operators and hops in local subgraphs, respectively.  

\vskip 1.5mm
\noindent \textbf{\pos vs. \sop.} \pos and \sop are similar in capturing information diffusion among nodes (at most) $r$-hop apart. But, their key distinction is whether long-range diffusion occurs in the global level of the input graph (for \sop) or the local level of the subgraph (for \pos). \sop is not a ``typical'' SGRL, however, it still uses subgraphs around the target pair on the power of graphs.

\subsection{Graph Pooling}
Our proposed instances of \ssgrl are completed by defining the graph pooling function in Eq.~\ref{eq:simple_sgrl}. We specifically consider resource-efficient pooling functions, which focus on pooling the targets' or their direct neighbors' learned representations.\footnote{Our focus is backed up by recent findings \cite{chamberlain2023graph} showing the effectiveness of the target node's embeddings and the decline in the informativeness of node embeddings as the nodes get farther away from targets.} We consider simple \textit{center} pooling: $$\pool_C(\mathbf{Z})=\mathbf{z}_u \odot \mathbf{z}_v,$$ where $\odot$ is the Hadamard product, and $\mathbf{z}_{u}$, $\mathbf{z}_{v}$ are $u$'s and $v$'s learned representations in nodal-representation matrix $\mathbf{Z}$
. We also introduce \textit{common-neighbor} pooling: $$\pool_{N}(\mathbf{Z})= AGG(\{\mathbf{z}_{i} | i \in N_u\cap N_v\}),$$ where $AGG$ is any invariant graph readout function (e.g., mean, max, or sum), and $N_u$ and $N_v$ are the direct neighborhood of targets $u$, $v$ in the original input graph. We also define \textit{center-common-neighbor} pooling $$\pool_{CCN}=\pool_C \oplus \pool_N,$$ where $\oplus$ is the concatenation operator. In addition to their efficacy, these pooling functions allow one to further optimize the data storage and computations. As the locations of \textit{pooled nodes} (e.g., target pairs and/or their direct neighbors) for these pooling functions are specified in the original graph, one could just conduct necessary (pre)computations and store those data affecting the learned representation of pooled nodes, while avoiding unnecessary computations and data storage. Figure \ref{fig:architecture} shows this optimization through red-bordered boxes of rows for each $\zruv$. This information is the only required node representations in $\zruv$ utilized in the downstream $\pool$ operation. We consider center pooling as the default pooling for \pos and \sop, and we refer to them as \posplus and \sopplus when center-common-neighbor pooling is deployed. We study both center and center-common-neighbor pooling for \pos. However, due to the explosive growth of $h$-hop subgraph sizes in \sopplus, we limit our experiments on \sop to only center pooling.

\begin{table}[t]
\centering
\caption{The statistics of the non-attributed, attributed, and OGB datasets.}
\label{table:dataset_comparison}
\scalebox{0.92}{
\begin{tabular}{ll>{\centering}p{1cm}>{\centering}p{2cm}>{\centering\arraybackslash}p{1cm}l}  
\toprule
                                & \textbf{Dataset} & \multicolumn{1}{c}{\textbf{\# Nodes}} & \multicolumn{1}{c}{\textbf{\# Edges}} & \multicolumn{1}{c}{\textbf{Avg. Deg.}} & \textbf{\# Feat.}  \\ 
\midrule
\multirow{3}{*}{\begin{sideways}{\scriptsize Non-attr.}\end{sideways}} & \textrm{NS}               & 1,461                                  & 2,742                                  & 3.75                                   & NA                    \\
                                & \textrm{Power}            & 4,941                                  & 6,594                                & 2.67                                   & NA                    \\
                                & \textrm{Yeast}            & 2,375                                  & 11,693                                 & 9.85                                   & NA                    \\
                                
                                & \textrm{PB}               & 1,222                                  & 16,714                                 & 27.36                                  & NA                    \\ 
\cmidrule[0.75pt](r){2-6}
\multirow{4}{*}{\begin{sideways}{\scriptsize Attributed}\end{sideways}}     & \textrm{Cora}             & 2,708                                  & 4,488                                  & 3.31                                   & 1,433                  \\
                                & \textrm{CiteSeer}         & 3,327                                  & 3,870                                  & 2.33                                   & 3,703                  \\
                                & \textrm{PubMed}           & 19,717                                 & 37,676                                & 3.82                                   & 500                   \\
                                & \textrm{Texas}        & 183                                   & 143                                   & 1.56                                   & 1,703                  \\ 
                                & \textrm{Wisconsin}        & 251                                   & 197                                   & 1.57                                   & 1,703                  \\ 
\cmidrule[0.75pt](r){2-6}
\multirow{4}{*}{\begin{sideways}{\scriptsize OGB}\end{sideways}}            & \textrm{Collab}           & 235,868                                & 1,285,465                               & 8.2                                     & 128                   \\
& \textrm{DDI}           & 4,267                                & 1,334,889                               & 500.5                                     & NA                   \\
                                & \textrm{Vessel}           & 3,538,495                               & 5,345,897                               & 2.4                                   & 3                     \\
                                & \textrm{PPA}           &      576,289                          & 30,326,273 	                               & 73.7                                    & 58                     \\
                                & \textrm{Citation2}           & 2,927,963                               & 30,561,187                               & 20.7                                   & 128                     \\
\bottomrule
\end{tabular}
}
\vspace{-9pt}
\end{table}

\subsection{Time Complexities and Storage Requirement}\label{sec:time_Space}
We first discuss the (amortized) inference time complexity for a target pair (i.e., a single link) under any variants of our S3GRL. Then, we dive deeper and compare the preprocessing and inference time complexity of the \pos model and its competitive counterpart SEAL \cite{zhang2018link}. Table \ref{tab:time_complexity} summarizes our discussion. Finally, we discuss on the disk space requirement for prepossessing of \pos vs. SEAL. 
\vskip 1.5mm
\noindent \textbf{Inference Time Complexity.} Let $p$ be the number of pooled nodes (e.g., $p=2$ for center pooling), $d$ be the dimension of initial input features, $r$ be the number of operators, and $d'$ be the reduced dimensionality in Eq.~\ref{eq:simple_sgrl}. The inference time complexity for any of \pos, \sop, and their variants is $O(rpdd'+d'^2)$. Consider the dimensionality reduction of the joint matrix $\mathbf{Z}_{uv}$ by the weight matrix $\mathbf{W}$ in Eq.~\ref{eq:simple_sgrl}. As $\mathbf{Z}_{uv}$ is $rp$ by $d$ and $\mathbf{W}$ is $d$ by $d'$, their multiplication is in $O(rpdd')$ time. The pooling for any proposed variants is $O(pd')$. Assuming the transformation function $\Omega$ in Eq.~\ref{eq:link-prob-s3grl} being an MLP with one $d'$-dimensional hidden layer, its computation is in $O(d'^2)$ time.

\vskip 1.5mm
\noindent \textbf{Preprocessing time complexity (\pos vs SEAL).} Denoting the node feature dimension $d$, the number of operators $r$, the number of nodes $n$ and the number of edges $m = |E|$, the preprocessing time complexity of \pos is $O(rnm + rmd)$. The first term comes from the computation of the powers of sparse subgraph adjacency matrices, which are diffusion matrices in \pos. The second is for the application of diffusion matrices over nodal feature matrices to compute only the smoothed target pair's representations. We assumed that the subgraph adjacency matrices and their small powers are sparse and have $O(m)$ nonzero elements. To reduce this time complexity, one can control the subgraph size by sampling subgraphs with controllable parameters. For example, we can consider \pos+ ScaLed with diffusion operators of \pos operating on sampled enclosing subgraphs, where each subgraph is sampled by $k$ many $l$-length random walks rooted at each target node. One can notice that \pos+ ScaLed has the preprocessing time of $O(rk^2l^2+rkld)$ as the number of nodes in the subgraph is bounded above by $lk$. Note that $l$ and $k$ are set such that $lk \ll n$. For SEAL, we consider its static mode (each subgraph is generated and labeled as a preprocessing step) and dynamic mode (each subgraph is generated on the fly), with $O(m)$ and $O(1)$ preprocessing time, respectively.\footnote{The labelling and $h$-hop subgraph extraction require at least a traversal of the subgraph, which is $O(m)$ for complex networks \cite{chamberlain2023graph}, mainly due to their power-law degree distribution.}

\vskip 1.5mm
\noindent \textbf{Inference time complexity (\pos vs. SEAL).} Letting $d'$ be the dimensionality of hidden layers, the inference time complexity of \pos and its variants (e.g., \pos+ ScaLed) is $O(rdd'+d'^2)$. This is easily derived from the general inference time complexity discussed above by setting the number of pooled nodes $p=2$ for \pos. Note that this time complexity is independent of the subgraph size, the choice of subgraph samplings, and the choice of diffusion operators. For SEAL, regardless of its mode, the inference time complexity is $O(Lmdd'+d'^2)$, where $L$ is the number of convolution layers in DGCNN \cite{zhang2018end}. The first term corresponds to graph convolution operation and message passing over the subgraph. 
The second term is for MLP in SEAL. Note that the complexity of SEAL heavily depends on the subgraph size (i.e., number of edges $m$
). Comparing the time complexities of SEAL and \pos, $O(rdd')$ is noticeably faster than $O(Lmdd')$ as the number of the operator $r$ is considerably lower than the number edges $m$ (i.e., $r \ll m$).

\vskip 1.5mm
\noindent \textbf{Time Complexity Discussion (\pos vs SEAL).} The preprocessing complexity of \pos $O(rnm + rmd)$ is noticeably greater than that of SEAL for both dynamic $O(1)$ and static $O(m)$ modes . However, this higher complexity is manageable in practice by resorting to sampling such as \pos+ ScaLed $O(rk^2l^2+kld)$ or by leveraging distributed computing infrastructures such as Apache Spark. Even without such mitigation strategies, we argue that incurring this relatively high pre-processing cost to gain faster inference and training offers efficiency in practice. This contrasts SEAL, suffering from a considerably high inference time complexity of $O(Lmdd'+d'^2)$, which depends on graph size (e.g., the number of edges $m$). It is important to note that the computational cost of inference is repetitive and accumulated over multiple training iterations. Our experiments below validate this argument.

However, for single-instance inference (e.g., test data), S3GRL's scalability benefits may be less pronounced than SEAL, as both preprocessing and inference are required. In this scenario, the total complexity for \pos becomes $O(rnm + rmd + rdd' + d'^2)$, while for SEAL it is $O(Lmdd' + d'^2)$. Depending on the graph size ($n$ nodes and $m$ edges) and the feature dimensionalities ($d$ and $d'$), either method may be faster. Nonetheless, this does not undermine the scalability advantages of S3GRL over SEAL during training.

\vskip 1.5mm
\noindent \textbf{Disk Space Requirement (SEAL vs S3GRL).} The preprocessing in SGRLs imposes additional disk space requirements to store preprocessed datasets including enclosing subgraphs for each target pair, and their augmented nodal features. For SEAL, each target pair requires $O(m+nd)$ disk space for storing the subgraph with $m$ edges, $n$ nodes, and $d$ dimensional feature nodes. However, our S3GRl reduces this requirement substantially to $O(rpd)$, where $r$ is the number of diffusion operators and $p$ is the number of selected nodes for pooling (e.g., $p=2$ for \pos). Most importantly, the disk space requirement in S3GRL is independent of graph size as opposed to SEAL. Our experiments demonstrate that our models can achieve up to a 99\% data storage reduction compared to SEAL.   

\section{Experiments}
We carefully design an extensive set of experiments to assess the extent to which our model scales up SGRLs while maintaining their state-of-the-art performance. 
Our experiments intend to address these questions: \textbf{(Q1)} How effective is the \ssgrl framework compared with the state-of-the-art SGRLs for link prediction methods? \textbf{(Q2)} What is the computational gain achieved through \ssgrl in comparison to SGRLs? \textbf{(Q3)} How well does our best-performing model, \posplus, perform on the Open Graph Benchmark datasets for link prediction, graphs with millions of nodes and edges \cite{hu2020ogb,hu2021ogb}? \textbf{(Q4)} How complementary can \ssgrl be in combination with other scalable SGRLs (e.g., \scaled \cite{louis2022sampling}) to further boost scalability and generalization?

\begin{table*}[t]
\centering
\caption{Average AUC for attributed and non-attributed datasets (over 10 runs). The top 3 models are \textbf{\first{First}}, \textbf{\second{Second}}, and \textbf{\third{Third}}. \colorbox[rgb]{0.851,0.918,0.827}{Green} is best model among our \ssgrl variants. \colorbox[rgb]{1,0.949,0.8}{Yellow} is the best baseline. Gain is AUC difference of \colorbox[rgb]{0.851,0.918,0.827}{Green} and \colorbox[rgb]{1,0.949,0.8}{Yellow}.}
\label{auc-values}
\addtolength{\extrarowheight}{\belowrulesep}

\scalebox{0.96}{
\begin{tabular}{p{12pt}|l|llll|lllll} 
\toprule
\multicolumn{1}{l|}{}                                                    & \multirow{2}{*}{\textbf{Model}} & \multicolumn{4}{c|}{\textbf{Non-attributed}}                                                                                                                                                                       & \multicolumn{5}{c}{\textbf{Attributed}}                                                                                                                                                                                                                                \\
\multicolumn{1}{l|}{}                                                    &                        & \multicolumn{1}{c}{\textbf{NS}}                           & \multicolumn{1}{c}{\textbf{Power}}                        & \multicolumn{1}{c}{\textbf{PB}}                           & \multicolumn{1}{c|}{\textbf{Yeast}}                    & \multicolumn{1}{c}{\textbf{Cora}}                         & \multicolumn{1}{c}{\textbf{CiteSeer}}                     & \multicolumn{1}{c}{\textbf{PubMed}}                       & \multicolumn{1}{c}{\textbf{Texas}}                        & \multicolumn{1}{c}{\textbf{Wisconsin}}                     \\ 
\hline
\multirow{3}{*}{{\begin{sideways}Heuristic\end{sideways}}}                                        &  AA                     & 92.14{\scriptsize$\pm$0.77}                                     & 58.09{\scriptsize$\pm$0.55}                                     & 91.76{\scriptsize$\pm$0.56}                                     & 88.80{\scriptsize$\pm$0.55}                                     & 71.48{\scriptsize$\pm$0.69}                                     & 65.86{\scriptsize$\pm$0.80}                                     & 64.26{\scriptsize$\pm$0.40}                                     & 54.69{\scriptsize$\pm$3.68}                                     & 55.60{\scriptsize$\pm$3.14}                                      \\
            & CN                     & 92.12{\scriptsize$\pm$0.79}                                     & 58.09{\scriptsize$\pm$0.55}                                     & 91.44{\scriptsize$\pm$0.59}                                     & 88.73{\scriptsize$\pm$0.56}                                     & 71.40{\scriptsize$\pm$0.69}                                     & 65.84{\scriptsize$\pm$0.81}                                     & 64.26{\scriptsize$\pm$0.40}                                     & 54.36{\scriptsize$\pm$3.65}                                     & 55.08{\scriptsize$\pm$3.08}                                      \\
    & PPR                    & 92.50{\scriptsize$\pm$1.06}                                     & 62.88{\scriptsize$\pm$2.18}                                     & 86.85{\scriptsize$\pm$0.48}                                     & 91.71{\scriptsize$\pm$0.74}                                     & 82.87{\scriptsize$\pm$1.01}                                     & 74.35{\scriptsize$\pm$1.51}                                     & 75.80{\scriptsize$\pm$0.35}                                     & 53.81{\scriptsize$\pm$7.53}                                     & 62.86{\scriptsize$\pm$8.13}                                      \\ 
\hline
\multirow{3}{*}{{\begin{sideways}MPGNN\end{sideways}}}                                               & GCN                    & 91.75{\scriptsize$\pm$1.68}                                     & 69.41{\scriptsize$\pm$0.90}                                     & 90.80{\scriptsize$\pm$0.43}                                     & 91.29{\scriptsize$\pm$1.11}                                     & 89.14{\scriptsize$\pm$1.20}                                     & 87.89{\scriptsize$\pm$1.48}                                     & 92.72{\scriptsize$\pm$0.64}                                     & 67.42{\scriptsize$\pm$9.39}                                     & 72.77{\scriptsize$\pm$6.96}                                      \\
            & GraphSAGE              & 91.39{\scriptsize$\pm$1.73}                                     & 64.94{\scriptsize$\pm$2.10}                                     & 88.47{\scriptsize$\pm$2.56}                                     & 87.41{\scriptsize$\pm$1.64}                                     & 85.96{\scriptsize$\pm$2.04}                                     & 84.05{\scriptsize$\pm$1.72}                                     & 81.60{\scriptsize$\pm$1.22}                                     & 53.59{\scriptsize$\pm$9.37}                                     & 61.81{\scriptsize$\pm$9.66}                                      \\
             & GIN                    & 83.26{\scriptsize$\pm$3.81}                                     & 58.28{\scriptsize$\pm$2.61}                                     & 88.42{\scriptsize$\pm$2.09}                                     & 84.00{\scriptsize$\pm$1.94}                                     & 68.74{\scriptsize$\pm$2.74}                                     & 69.63{\scriptsize$\pm$2.77}                                     & 82.49{\scriptsize$\pm$2.89}                                     & 63.46{\scriptsize$\pm$8.87}                                     & 70.82{\scriptsize$\pm$8.25}                                      \\ 
\hline
\multirow{2}{*}{{\begin{sideways}LF\end{sideways}}} & node2vec               & 91.44{\scriptsize$\pm$0.81}                                     & 73.02{\scriptsize$\pm$1.32}                                     & 85.08{\scriptsize$\pm$0.74}                                     & 90.60{\scriptsize$\pm$0.57}                                     & 78.32{\scriptsize$\pm$0.74}                                     & 75.36{\scriptsize$\pm$1.22}                                     & 79.98{\scriptsize$\pm$0.35}                                     & 52.81{\scriptsize$\pm$5.31}                                     & 59.57{\scriptsize$\pm$5.69}                                      \\
              & MF                     & 82.56{\scriptsize$\pm$5.90}                                     & 53.83{\scriptsize$\pm$1.76}                                     & 91.56{\scriptsize$\pm$0.56}                                     & 87.57{\scriptsize$\pm$1.64}                                     & 62.25{\scriptsize$\pm$2.21}                                     & 61.65{\scriptsize$\pm$3.80}                                     & 68.56{\scriptsize$\pm$12.13}                                    & 60.35{\scriptsize$\pm$5.62}                                     & 53.75{\scriptsize$\pm$9.00}                                      \\ 
\hline
\multirow{4}{*}{{\begin{sideways}AE\end{sideways}}}  & GAE                    & 92.50{\scriptsize$\pm$1.71}                                     & 68.17{\scriptsize$\pm$1.64}                                     & 91.52{\scriptsize$\pm$0.35}                                     & 93.13{\scriptsize$\pm$0.79}                                     & 90.21{\scriptsize$\pm$0.98}                                     & 88.42{\scriptsize$\pm$1.13}                                     & 94.53{\scriptsize$\pm$0.69}                                     & 68.67{\scriptsize$\pm$6.95}                                     & 75.10{\scriptsize$\pm$8.69}                                      \\
                       & VGAE                   & 91.83{\scriptsize$\pm$1.49}                                     & 66.23{\scriptsize$\pm$0.94}                                     & 91.19{\scriptsize$\pm$0.85}                                     & 90.19{\scriptsize$\pm$1.38}                                     & 92.17{\scriptsize$\pm$0.72}                                     & 90.24{\scriptsize$\pm$1.10}                                     & 92.14{\scriptsize$\pm$0.19}                                     & \textbf{\third{74.61{\scriptsize$\pm$8.61}}}                                     & 74.39{\scriptsize$\pm$8.39}                                      \\
                    & ARVGA                  & 92.16{\scriptsize$\pm$1.05}                                     & 66.26{\scriptsize$\pm$1.59}                                     & 90.98{\scriptsize$\pm$0.92}                                     & 90.25{\scriptsize$\pm$1.06}                                     & \base \textbf{\third{92.26{\scriptsize$\pm$0.74}}}       & 90.29{\scriptsize$\pm$1.01}                                     & 92.10{\scriptsize$\pm$0.38}                                     & 73.55{\scriptsize$\pm$9.01}                                     & 72.65{\scriptsize$\pm$7.02}                                      \\
                   & GIC                    & 90.88{\scriptsize$\pm$1.85}                                     & 62.01{\scriptsize$\pm$1.25}                                     & 73.65{\scriptsize$\pm$1.36}                                     & 88.78{\scriptsize$\pm$0.63}                                     & 91.42{\scriptsize$\pm$1.24}                                     & \base \textbf{\third{92.99{\scriptsize$\pm$1.14}}}       & 91.04{\scriptsize$\pm$0.61}                                     & 65.16{\scriptsize$\pm$7.87}                                     & 75.24{\scriptsize$\pm$8.45}                                      \\ 
\hline
\multirow{3}{*}{{\begin{sideways}SGRL\end{sideways}}}                                                   & SEAL                   & \textbf{\third{98.63{\scriptsize$\pm$0.67}}}                                     & 85.28{\scriptsize$\pm$0.91}                                     & \textbf{\third{95.07{\scriptsize$\pm$0.35}}}                                     & \textbf{\second{97.56{\scriptsize$\pm$0.32}}}                                     & 90.29{\scriptsize$\pm$1.89}                                     & 88.12{\scriptsize$\pm$0.85}                                     & 97.82{\scriptsize$\pm$0.28}                                     & 71.68{\scriptsize$\pm$6.85}                                     & 77.96{\scriptsize$\pm$10.37}                                     \\
                 & GCN+DE                 & \textbf{\second{98.66{\scriptsize$\pm$0.66}}}                                     & 80.65{\scriptsize$\pm$1.40}                                     & \textbf{\second{95.14{\scriptsize$\pm$0.35}}}                                     & 96.75{\scriptsize$\pm$0.41}                                     & 91.51{\scriptsize$\pm$1.10}                                     & 88.88{\scriptsize$\pm$1.53}                                     & 98.15{\scriptsize$\pm$0.11}                                     & \textbf{\second{76.60{\scriptsize$\pm$6.40}}}                                     & 74.65{\scriptsize$\pm$9.56}                                    \\
              & WalkPool               & \base \textbf{\first{98.92{\scriptsize$\pm$0.52}}}       & \base \textbf{\first{90.25{\scriptsize$\pm$0.64}}}       & \base \textbf{\first{95.50{\scriptsize$\pm$0.26}}}       & \base \textbf{\first{98.16{\scriptsize$\pm$0.20}}}       & 92.24{\scriptsize$\pm$0.65}                                     & 89.97{\scriptsize$\pm$1.01}                                     & \base \textbf{\third{98.36{\scriptsize$\pm$0.11}}}       & \base \textbf{\first{78.44{\scriptsize$\pm$9.83}}}       & \base\textbf{\second{79.57{\scriptsize$\pm$11.02}}}      \\ 
\hline
\multirow{3}{*}{{\begin{sideways}\ssgrl\end{sideways}}}                                                   & \pos (ours)                    & 97.23{\scriptsize$\pm$1.38}                                     & \textbf{\third{86.67{\scriptsize$\pm$0.98}}}                                     & 94.83{\scriptsize$\pm$0.41}                                     & 95.47{\scriptsize$\pm$0.54}                                     & \textbf{\second{94.65{\scriptsize$\pm$0.67}}}                                     & \best \textbf{\first{95.76{\scriptsize$\pm$0.59}}} & \textbf{\second{98.97{\scriptsize$\pm$0.08}}}                                     & 73.75{\scriptsize$\pm$8.20}                                     & \best \textbf{\first{82.50{\scriptsize$\pm$5.83}}}  \\
             & \posplus (ours)                  & \best 98.37{\scriptsize$\pm$1.26} & \best \textbf{\second{87.82{\scriptsize$\pm$0.96}}} & \best 95.04{\scriptsize$\pm$0.27} & \best \textbf{\third{96.77{\scriptsize$\pm$0.39}}} & \best \textbf{\first{94.77{\scriptsize$\pm$0.68}}} & \textbf{\second{95.72{\scriptsize$\pm$0.56}}}                                     & \best \textbf{\first{99.00{\scriptsize$\pm$0.08}}} & \best \textbf{\first{78.44{\scriptsize$\pm$9.83}}} & \textbf{\third{79.17{\scriptsize$\pm$10.87}}}                                     \\
            & \sop (ours)                    & 90.61{\scriptsize$\pm$1.94}                                     & 75.64{\scriptsize$\pm$1.33}                                     & 94.34{\scriptsize$\pm$0.30}                                     & 92.98{\scriptsize$\pm$0.58}                                     & 91.24{\scriptsize$\pm$0.80}                                     & 88.23{\scriptsize$\pm$0.73}                                     & 95.91{\scriptsize$\pm$0.29}                                     & 69.49{\scriptsize$\pm$7.12}                                     & 72.29{\scriptsize$\pm$14.42}                                     \\ 
\hline
\multicolumn{2}{c|}{Gain}                                                                          & \multicolumn{1}{c}{-0.55}                        & \multicolumn{1}{c}{-2.43}                        & \multicolumn{1}{c}{-0.46}                        & \multicolumn{1}{c|}{-1.39}                       & \multicolumn{1}{c}{+2.51}                         & \multicolumn{1}{c}{+2.77}                         & \multicolumn{1}{c}{+0.64}                         & \multicolumn{1}{c}{0}                            & \multicolumn{1}{c}{+2.93}                          \\
\bottomrule
\end{tabular}
}
\end{table*}

\begin{table*}[tb]
\centering
\setlength{\extrarowheight}{0pt}
\addtolength{\extrarowheight}{\aboverulesep}
\addtolength{\extrarowheight}{\belowrulesep}
\setlength{\aboverulesep}{0pt}
\setlength{\belowrulesep}{0pt}
\arrayrulecolor{black}
\caption{Computation time of SGRLs vs. our \ssgrl models: average training time (over 50 epochs), average inference time, preprocessing time, and total runtime (preprocessing, training, and inference time) for 50 epochs. \colorbox[rgb]{0.851,0.918,0.827}{Green} is the fastest and \colorbox[rgb]{0.957,0.8,0.8}{Red} is slowest for each group of SGRLs and \ssgrl. Max(min) speedup corresponds to the ratio of time taken by the slowest (fastest) SGRLs to our fastest (slowest) model.}
\label{resource-cons}
\scalebox{0.81}{
\begin{tabular}{l|cccc|cccc|cccc} 
\toprule
\textbf{Model}              & \textbf{Training}                                 & \textbf{Inference}                              & \textbf{Preproc.}                          & \textbf{Runtime}                            & \textbf{Training}                                & \textbf{Inference}                              & \textbf{Preproc.}                         & \textbf{Runtime}                           & \textbf{Training}                                & \textbf{Inference}                              & \textbf{Preproc.}                          & \textbf{Runtime}                             \\ 
\hline
\multicolumn{1}{l|}{}       & \multicolumn{4}{c|}{\textbf{NS }(non-attributed)}                                                                                                                                              & \multicolumn{4}{c|}{\textbf{Power }(non-attributed)}                                                                                                                                        & \multicolumn{4}{c}{\textbf{Yeast }(non-attributed)}                                                                                                                                            \\ 
\hline
SEAL                        & 4.91{\scriptsize$\pm$0.23}                                       & 0.14{\scriptsize$\pm$0.01}                                     & \worst17.86      & 275.28                                      & 11.73{\scriptsize$\pm$0.02}                                     & 0.33{\scriptsize$\pm$0.01}                                     & \worst44.48     & 658.14                                     & 24.03{\scriptsize$\pm$0.40}                                     & 0.54{\scriptsize$\pm$0.05}                                     & \worst115.02     & 1362.85                                      \\
GCN+DE                      & \best3.58{\scriptsize$\pm$0.12}   & \best0.10{\scriptsize$\pm$0.01} & \best11.73  & \best198.98  & \best8.62{\scriptsize$\pm$0.27}  & \best0.25{\scriptsize$\pm$0.01} & \best28.59 & \best479.4  & \best18.41{\scriptsize$\pm$0.71} & \best0.46{\scriptsize$\pm$0.06} & \best82.19  & \best1040.72  \\
WalkPool                    & \worst7.66{\scriptsize$\pm$0.09}       & \worst0.41{\scriptsize$\pm$0.02}     & 12.18                                      & \worst427.03      & \worst18.46{\scriptsize$\pm$0.76}     & \worst0.87{\scriptsize$\pm$0.06}     & 33.51                                     & \worst1024.55    & \worst174.80{\scriptsize$\pm$1.06}    & \worst8.05{\scriptsize$\pm$0.11}     & 90.75                                      & \worst9443.17      \\ 
\hline
\pos                        & \best2.24{\scriptsize$\pm$0.15}   & \best0.06{\scriptsize$\pm$0.01} & 34.86                                      & 152.23                                      & 5.58{\scriptsize$\pm$0.48}                                      & \best0.14{\scriptsize$\pm$0.01} & 97.71                                     & 388.97                                     & 9.95{\scriptsize$\pm$1.45}                                      & 0.20{\scriptsize$\pm$0.06}                                     & \worst259.96     & \worst775.58       \\
\posplus                       & \worst2.54{\scriptsize$\pm$0.06}       & \worst0.07{\scriptsize$\pm$0.00}     & \worst41.43      & \worst173.78      & \worst6.12{\scriptsize$\pm$0.24}      & \worst0.16{\scriptsize$\pm$0.01}     & \worst107.77    & \worst426.53     & \worst10.49{\scriptsize$\pm$0.61}     & \best0.20{\scriptsize$\pm$0.04} & 206.52                                     & 749.87                                       \\
\sop                         & 2.26{\scriptsize$\pm$0.11}                                       & 0.06{\scriptsize$\pm$0.00}                                     & \best24.67  & \best142.45  & \best5.41{\scriptsize$\pm$0.23}  & 0.14{\scriptsize$\pm$0.01}                                     & \best65.65 & \best347.62 & \best9.24{\scriptsize$\pm$0.74}  & \worst0.22{\scriptsize$\pm$0.04}     & \best117.23 & \best597.29   \\ 
\hline
\multicolumn{1}{c}{Speedup} & 3.42(1.41)                                        & 6.83(1.43)                                      & 0.72(0.28)                                 & \multicolumn{1}{c}{3.00(1.14)}                 & 3.41(1.41)                                       & 6.21(1.56)                                      & 0.68(0.27)                                & \multicolumn{1}{c}{2.95(1.12)}             & 18.92(1.76)                                      & 40.25(2.09)                                     & 0.98(0.32)                                 & 15.81(1.34)                                  \\ 
\hline
\multicolumn{1}{l|}{}       & \multicolumn{4}{c|}{\textbf{PB }(non-attributed)}                                                                                                                                              & \multicolumn{4}{c|}{\textbf{Cora }(attributed)}                                                                                                                                             & \multicolumn{4}{c}{\textbf{CiteSeer }(attributed)}                                                                                                                                             \\ 
\hline
SEAL                        & 64.62{\scriptsize$\pm$5.59}                                      & 2.32{\scriptsize$\pm$0.10}                                     & \worst531.79     & 3947.45                                     & 18.37{\scriptsize$\pm$1.49}                                     & 0.73{\scriptsize$\pm$0.12}                                     & \worst113.32    & \worst1090.94    & 12.54{\scriptsize$\pm$0.69}                                     & 0.58{\scriptsize$\pm$0.10}                                     & \worst93.52      & 768.72                                       \\
GCN+DE                      & \best55.82{\scriptsize$\pm$1.59}  & \best2.01{\scriptsize$\pm$0.09} & 398.81                                     & \best3346.80 & \best14.85{\scriptsize$\pm$0.53} & \best0.62{\scriptsize$\pm$0.08} & 80.48                                     & \best872.68 & \best11.43{\scriptsize$\pm$0.71} & \best0.52{\scriptsize$\pm$0.07} & 71.97                                      & \best685.98   \\
WalkPool                    & \worst133.30{\scriptsize$\pm$0.52}     & \worst6.48{\scriptsize$\pm$0.15}     & \best136.29 & \worst7291.50     & \worst18.53{\scriptsize$\pm$0.91}     & \worst1.00{\scriptsize$\pm$0.15}     & \best27.43 & 1034.33                                    & \worst15.32{\scriptsize$\pm$0.54}     & \worst0.87{\scriptsize$\pm$0.05}     & \best22.82  & \worst859.27       \\ 
\hline
\pos                         & 13.42{\scriptsize$\pm$0.77}                                      & \worst0.33{\scriptsize$\pm$0.04}     & 1754.88                                    & 2452.39                                     & 5.44{\scriptsize$\pm$0.52}                                      & \best0.15{\scriptsize$\pm$0.02} & \worst106.45    & 394.12                                     & \best4.82{\scriptsize$\pm$0.22}  & \best0.15{\scriptsize$\pm$0.01} & \worst78.62      & \worst335.37       \\
\posplus                       & \worst15.56{\scriptsize$\pm$1.28}      & 0.29{\scriptsize$\pm$0.05}                                     & \worst2527.23    & \worst3331.56     & \worst5.87{\scriptsize$\pm$0.17}      & \worst0.17{\scriptsize$\pm$0.01}     & 93.69                                     & \worst401.05     & 4.91{\scriptsize$\pm$0.39}                                      & \worst0.17{\scriptsize$\pm$0.02}     & 72.02                                      & 331.55                                       \\
\sop                         & \best13.32{\scriptsize$\pm$0.72}  & \best0.25{\scriptsize$\pm$0.06} & \best333.58 & \best1022.90 & \best5.04{\scriptsize$\pm$0.17}  & 0.15{\scriptsize$\pm$0.03}                                     & \best35.65 & \best300.10 & \worst4.96{\scriptsize$\pm$0.14}      & 0.17{\scriptsize$\pm$0.01}                                     & \best31.57  & \best293.96   \\ 
\hline
\multicolumn{1}{c}{Speedup} & 10.01(3.59)                                       & 25.92(6.09)                                     & 1.59(0.05)                                 & \multicolumn{1}{c}{7.13(1.00)}              & 3.68(2.53)                                       & 6.67(3.65)                                      & 3.18(0.26)                                & \multicolumn{1}{c}{3.64(2.18)}             & 3.18(2.30)                                        & 5.80(3.06)                                       & 2.96(0.29)                                 & 2.92(2.05)                                   \\ 
\hline
\multicolumn{1}{l|}{}       & \multicolumn{4}{c|}{\textbf{Pubmed }(attributed)}                                                                                                                                              & \multicolumn{4}{c|}{\textbf{Texas }(attributed)}                                                                                                                                            & \multicolumn{4}{c}{\textbf{Wisconsin }(attributed)}                                                                                                                                            \\ 
\hline
SEAL                        & \worst533.18{\scriptsize$\pm$4.64}     & \worst38.46{\scriptsize$\pm$1.08}    & 141.76                                     & \worst30150.31    & 0.32{\scriptsize$\pm$0.01}                                      & \best0.01{\scriptsize$\pm$0.00} & \worst2.55      & 20.46                                      & 0.47{\scriptsize$\pm$0.01}                                      & \best0.02{\scriptsize$\pm$0.00} & \worst3.29       & 29.27                                        \\
GCN+DE                      & 423.73{\scriptsize$\pm$2.67}                                     & 34.44{\scriptsize$\pm$1.21}                                    & \best106.00 & 24311.00                                    & \best0.31{\scriptsize$\pm$0.01}  & 0.01{\scriptsize$\pm$0.00}                                     & 1.87                                      & \best18.55  & \best0.43{\scriptsize$\pm$0.01}  & 0.02{\scriptsize$\pm$0.00}                                     & 2.63                                       & \best26.19    \\
WalkPool                    & \best150.27{\scriptsize$\pm$6.22} & \best8.10{\scriptsize$\pm$1.06} & \worst341.12     & \best8474.72 & \worst0.55{\scriptsize$\pm$0.08}      & \worst0.03{\scriptsize$\pm$0.01}     & \best0.92  & \worst32.54      & \worst0.85{\scriptsize$\pm$0.04}      & \worst0.06{\scriptsize$\pm$0.00}     & \best1.08   & \worst49.38        \\ 
\hline
\pos                         & 38.90{\scriptsize$\pm$2.89}                                      & 0.79{\scriptsize$\pm$0.10}                                     & \worst2986.74    & 5017.78                                     & 0.16{\scriptsize$\pm$0.01}                                      & \best0.01{\scriptsize$\pm$0.00} & 1.87                                      & \worst12.71      & \best0.21{\scriptsize$\pm$0.02}  & \best0.01{\scriptsize$\pm$0.00} & 2.65                                       & 15.86                                        \\
\posplus                       & \worst45.32{\scriptsize$\pm$2.21}      & \worst0.92{\scriptsize$\pm$0.11}     & 2976.80                                    & \worst5335.86     & \worst0.18{\scriptsize$\pm$0.01}      & 0.01{\scriptsize$\pm$0.00}                                     & \worst2.26      & 11.96                                      & \worst0.24{\scriptsize$\pm$0.02}      & 0.01{\scriptsize$\pm$0.00}                                     & \worst2.91       & \worst16.07        \\
\sop                         & \best38.38{\scriptsize$\pm$2.90}  & \best0.75{\scriptsize$\pm$0.15} & \best526.62 & \best2526.22 & \best0.15{\scriptsize$\pm$0.01}  & \worst0.01{\scriptsize$\pm$0.00}     & \best1.34  & \best9.61   & 0.22{\scriptsize$\pm$0.01}                                      & \worst0.01{\scriptsize$\pm$0.00}     & \best1.87   & \best13.75    \\ 
\hline
\multicolumn{1}{c}{Speedup} & 13.89(3.32)                                       & 51.28(8.80)                                     & 0.65(0.04)                                 & \multicolumn{1}{c}{11.93(1.59)}             & 2.62(1.72)                                       & 3.00(1.00)                                            & 1.90(0.41)                                 & \multicolumn{1}{c}{3.39(1.46)}             & 4.05(1.79)                                       & 6.00(2.00)                                            & 1.76(0.37)                                 & 3.59(1.63)                                   \\
\bottomrule
\end{tabular}
}
\end{table*}
\vskip 1.5mm
\noindent \textbf{Dataset.} For our experiments, we use undirected, weighted and unweighted, attributed and non-attributed datasets that are publicly available and commonly used in other link prediction studies \cite{zhang2018link,zhang2018end,li2020distance,pan2022neural,louis2022sampling,chamberlain2023graph}. Table \ref{table:dataset_comparison} shows the statistics of our datasets. We divide our datasets into three categories: non-attributed datasets, attributed datasets and the OGB datasets. The edges in the attributed and non-attributed dataset categories are randomly split into 85\% training, 5\% validation, and 10\% testing sets, except for Cora, CiteSeer and Pubmed datasets chosen for comparison in Table \ref{table:ogb-results}, where we follow the experimental setup in \cite{chamberlain2023graph} with a random split of 70\% training, 10\% validation, and 20\% testing sets. For the OGB datasets, we follow the official dataset split \cite{hu2021ogb}.

\vskip 1.5mm \noindent \textbf{Baselines.} To thoroughly evaluate the performance of our \ssgrl models (\pos, \posplus, and \sop), we compare them with a diverse set of 18 baseline methods spanning five categories of link prediction models. The selection of baselines is motivated by the need to cover various techniques, including classic and recent state-of-the-art approaches, to compare comprehensively. Our \textit{heuristic} benchmarks include common neighbors (CN), Adamic Adar (AA) \cite{adamic2003friends}, and Personalized PageRank (PPR), which are traditional yet effective and scalable methods for link prediction. For \textit{message-passing GNNs (MPGNNs)}, we select widely used architectures such as GCN \cite{kipf2017semi}, GraphSAGE \cite{hamilton2017inductive}, and GIN \cite{xu2019GIN}. These models represent foundational MPGNN techniques commonly used as benchmarks in the literature. Our \textit{latent factor (LF)} benchmarks are node2vec \cite{grover2016node2vec} and Matrix Factorization (MF) \cite{koren2009matrix}, which are representative of embedding-based methods for link prediction. The \textit{autoencoder (AE)} methods include GAE and VGAE \cite{kipf2016variational}, adversarially regularized variational graph autoencoder (ARVGA) \cite{pan2018adversarially}, and Graph InfoClust (GIC) \cite{Mavromatis2021GraphIM}. These methods are selected for their ability to learn effective representations through unsupervised learning. To ensure inclusion of recent advancements in subgraph representation learning (SGRL), our examined SGRLs include SEAL \cite{zhang2018link}, GCN+DE \cite{li2020distance}, WalkPool \cite{pan2022neural}, and BUDDY \cite{chamberlain2023graph}. SEAL is a seminal work in SGRL for link prediction, while GCN+DE incorporates distance encoding for enhanced performance. WalkPool and BUDDY are more recent methods that have shown strong performance on large-scale graphs. For OGB datasets, we compare against BUDDY \cite{chamberlain2023graph} and its baselines (i.e., a subset of our baselines listed above). By selecting these baselines, we aim to provide a comprehensive evaluation of our models against both traditional and cutting-edge methods, addressing the need for up-to-date comparisons.

\vskip 1.5mm
\noindent \textbf{Setup.} For SGRL and \ssgrl methods, we set the number of hops $h=2$ for the non-attributed datasets (except WalkPool on the Power dataset with $h=3$), $h=3$ on attributed datasets (except for WalkPool with $h=2$ based on \cite{pan2022neural}), and $h=1$ for OGB datasets. \footnote{All hyperparameters of baselines are optimally selected based on their original papers or the shared public implementations. When possible, we also select the \ssgrl hyperparameters to match the benchmarks' hyperparameters for a fair comparison.} In \ssgrl models, we set the number of operators $r=3$ for all datasets except ogbl-collab, ogbl-ppa and ogbl-citation2 where $r=1$. We also use the zero-one labeling scheme for all datasets except for ogbl-citation2 and ogbl-ppa, where DRNL is used instead. The $AGG$ graph readout function in center-common-neighbor pooling is set to a simple mean aggregation, except for ogbl-vessel, ogbl-ppa and ogbl-citation2 where sum is used instead. In \ssgrl models, across the attributed and non-attributed datasets, we set the hidden dimension in Eq.~\ref{eq:simple_sgrl} to $256$, and also implement $\Omega$ in Eq.~\ref{eq:link-prob-s3grl} as an MLP with one $256$-dimensional hidden layer. For all models, we set the dropout to 0.5 and train them for 50 epochs with Adam \cite{kingma2014adam} and a batch size of 32 (except for MPGNNs with full-batch training).



For the comparison against BUDDY, the results are taken from \cite{chamberlain2023graph}, except for ogbl-vessel dataset, where the baseline figures are taken from the publicly-shared leaderboards.\footnote{We report the baseline results as of April 2, 2023 from \href{https://ogb.stanford.edu/docs/leader\_linkprop}{https://ogb.stanford.edu/docs/leader\_linkprop}, except for BUDDY, run by adding support from \href{https://github.com/melifluos/subgraph-sketching}{https://github.com/melifluos/subgraph-sketching}.}



For all models, on the attributed and non-attributed datasets, we report the average of the area under the curve (AUC) of the testing data over 10 runs with different fixed random seeds.\footnote{We exclude Average Precision (AP) results due to their known strong correlations with AUC results \cite{pan2022neural}.} 
For the experiments against BUDDY on OGB and Planetoid datasets, we report the average of 10 runs (with different fixed random seeds) on the efficacy measures consistent with \cite{chamberlain2023graph}.

In each run, we test a model on the testing data with those parameters with the highest efficacy measure on the validation data. Our code is implemented in PyTorch Geometric \cite{Fey/Lenssen/2019} and PyTorch \cite{paszke2019pytorch}
.\footnote{Our code is available at \href{https://github.com/venomouscyanide/S3GRL\_OGB}{https://github.com/venomouscyanide/S3GRL\_OGB}. All experiments are run on servers with 50 CPU cores, 377 GB RAM, and 11 GB GTX 1080 Ti GPUs.} To compare the computational efficiency, we report the average training and inference time (over 50 epochs) and the total preprocessing and runtime for a fixed run on the attributed and non-attributed datasets. 



\begin{table*}
\addtolength{\extrarowheight}{\belowrulesep}
\centering
\caption[Results for \posplus in comparison to the methodology set in BUDDY \cite{chamberlain2023graph}]{Results for \posplus in comparison to the datasets and baselines chosen in BUDDY \cite{chamberlain2023graph}. The top three performing models are highlighted in \textbf{\first{First}}, \textbf{\second{Second}}, and \textbf{\third{Third}}.}
\label{table:ogb-results}
\scalebox{1.08}{
\begin{tabular}{l|c|c|c|c|c|c|c|c} 
\toprule
\textbf{Model} & \textbf{Cora} & \textbf{CiteSeer} & \textbf{Pubmed} & \textbf{Collab} & \textbf{DDI} & \textbf{Vessel}  & \textbf{Citation2} & \textbf{PPA} \\
               & HR@100        & HR@100            & HR@100          & HR@50           & HR@20        & roc-auc        & MRR  & HR@100 \\ 
\hline
CN             & 33.92{\scriptsize$\pm$0.46}    & 29.79{\scriptsize$\pm$0.90}        & 23.13{\scriptsize$\pm$0.15}      & 56.44{\scriptsize$\pm$0.00}      & 17.73{\scriptsize$\pm$0.00}   & 48.49{\scriptsize$\pm$0.00}  & 51.47{\scriptsize$\pm$0.00} & 27.65{\scriptsize$\pm$0.00} \\
AA             & 39.85{\scriptsize$\pm$1.34}    & 35.19{\scriptsize$\pm$1.33}        & 27.38{\scriptsize$\pm$0.11}      & 64.35{\scriptsize$\pm$0.00}      & 18.61{\scriptsize$\pm$0.00}   & 48.49{\scriptsize$\pm$0.00}  & 51.89{\scriptsize$\pm$0.00} & 32.45{\scriptsize$\pm$0.00} \\
GCN            & 66.79{\scriptsize$\pm$1.65}    & 67.08{\scriptsize$\pm$2.94}        & 53.02{\scriptsize$\pm$1.39}      & 44.75{\scriptsize$\pm$1.07}      & \textbf{\third{37.07{\scriptsize$\pm$5.07}}}   & 43.53{\scriptsize$\pm$9.61}    &  84.74{\scriptsize$\pm$0.21} & 18.67{\scriptsize$\pm$1.32} \\
GraphSAGE           & 55.02{\scriptsize$\pm$4.03}    & 57.01{\scriptsize$\pm$3.74}        & 39.66{\scriptsize$\pm$0.72}      & 48.10{\scriptsize$\pm$0.81}      & \textbf{\second{53.90{\scriptsize$\pm$4.74}}}   & 49.89{\scriptsize$\pm$6.78}   &  82.60{\scriptsize$\pm$0.36} & 16.55{\scriptsize$\pm$2.40} \\
SEAL           & \textbf{\third{81.71{\scriptsize$\pm$1.30}}}    & \textbf{\third{83.89{\scriptsize$\pm$2.15}}}        & \textbf{\second{75.54{\scriptsize$\pm$1.32}}}      & \textbf{\third{64.74{\scriptsize$\pm$0.43}}}      & 30.56{\scriptsize$\pm$3.86}   & \textbf{\second{80.50{\scriptsize$\pm$0.21}}}   &  \textbf{\second{87.67{\scriptsize$\pm$0.32}}}  & \textbf{\second{48.80{\scriptsize$\pm$3.16}}} \\
BUDDY          & \textbf{\second{88.00{\scriptsize$\pm$0.44}}}    & \textbf{\second{92.93{\scriptsize$\pm$0.27}}}        & \textbf{\third{74.10{\scriptsize$\pm$0.78}}}      & \textbf{\second{65.94{\scriptsize$\pm$0.58}}}      & \textbf{\first{78.51{\scriptsize$\pm$1.36}}}   & \textbf{\third{55.14{\scriptsize$\pm$0.11}}}          &   \textbf{\third{87.56{\scriptsize$\pm$0.11}}}   & \textbf{\first{49.85{\scriptsize$\pm$0.20}}} \\
\posplus (ours)           & \textbf{\first{91.55{\scriptsize$\pm$1.16}}}  & \textbf{\first{94.79{\scriptsize$\pm$0.58}}}      & \textbf{\first{79.40{\scriptsize$\pm$1.53}}}    & \textbf{\first{66.83{\scriptsize$\pm$0.30}}}    & 22.24{\scriptsize$\pm$3.36} & \textbf{\first{80.56{\scriptsize$\pm$0.06}}}   &  \textbf{\first{88.14{\scriptsize$\pm$0.08}}}  & \textbf{\third{42.42{\scriptsize$\pm$1.80}}} \\
\bottomrule
\end{tabular}
}
\end{table*}

\begin{table*}[tb]
    \centering
    \caption{Average preprocessing and inference runtimes (seconds) for a target pair in OGB datasets. SEAL-S and SEAL-D correspond to SEAL with the static and dynamic modes.}
\label{tab:time_OGB}
    \scalebox{1.0}{
    \begin{tabular}{lccccccccc}
        \toprule
        & \multicolumn{3}{c}{\textbf{Vessel}} & \multicolumn{3}{c}{\textbf{PPA}} & \multicolumn{3}{c}{\textbf{Citation2}} \\
        \cmidrule(lr){2-4} \cmidrule(lr){5-7} \cmidrule(lr){8-10}
        &  \posplus & SEAL-S& SEAL-D & \posplus & SEAL-S & SEAL-D&  \posplus & SEAL-S & SEAL-D\\
        \midrule
        Preprocessing  & 2.3E-02 & 1.9E-02 & - & 3.2E-02 & 1.8E-02 & - & 1.6E-02 & 1.0E-02 & - \\
        Inference  & 9.8E-05 & 4.0E-04 & 1.0E-03 & 1.2E-04 & 1.6E-03 & 1.3E-03 & 2.1E-04 & 3.4E-04 & 1.2E-03 \\
        \bottomrule
    \end{tabular}
    }
\end{table*}

\vskip 1.5mm
\noindent \textbf{Results: Attributed and Non-attributed Datasets.} On the attributed datasets, the \ssgrl models, particularly \posplus and \pos, consistently outperform others (see Table \ref{auc-values}). Their gain/improvement, compared to the best baseline, can reach $2.93$ (for Wisconsin) and $2.77$ (for CiteSeer). This state-of-the-art performance of \posplus and \pos suggests that our simplification techniques do not weaken SGRLs' efficacy and even improve their generalizability. Most importantly, this AUC gain is achieved by multi-fold less computation: The \ssgrl models benefit \grA{2.3--13.9} speedup in training time for citation network datasets of Cora, CiteSeer, and Pubmed (see Table \ref{resource-cons}). Similarly, inference time witnesses \grA{3.1--51.2} speedup, where the maximum speedup is achieved on the largest attributed dataset Pubmed. Our \ssgrl models exhibit higher dataset preprocessing times compared to SGRLs (see Table \ref{resource-cons}). However, this is easily negated by our models' faster accumulative training and inference times that lead to  \grA{1.4--11.9} overall runtime speedup across all attributed datasets (min. for Texas and max. for Pubmed). 

We observe comparatively lower AUC values for \sop, which can be attributed to longer-range information being of less value on these datasets. Regardless, \sop still shows comparable AUC to the other SGRLs while offering substantially higher speedups. 

Despite only consuming the source-target information, \pos achieves first or second place in the citation networks indicating the power of the center pooling. Moreover, the higher efficacy of \posplus compared to \pos across a few datasets indicate added expressiveness provided by center-common-neighbor pooling. Finally, the autoencoder methods outperform SGRLs on citation networks. However, \ssgrl instances outperform them and have enhanced the learning capability of SGRLs.

For the non-attributed datasets, although WalkPool outperforms others, we see strong performance from our \ssgrl instances. Our instances appear second or third (e.g., Yeast or Power) or have a small margin to the best model (e.g., NS or PB). The maximum loss in our models is bounded to 2.43\% (see Power's gain in Table \ref{auc-values}). Regardless of the small loss of efficacy, our \ssgrl models demonstrate multi-fold speedup in training and inference times: training with \grA{1.4--18.9} speedup and inference with \grA{1.4--40.2} (the maximum training and inference speedup on Yeast). We see a similar pattern of higher preprocessing times for our models; however, it gets negated by faster training and inference times leading to an overall speedup in runtimes with the maximum of \grA{15.8} for Yeast. \sop shows a relatively lower AUC in the non-attributed dataset except for PB and Yeast, possibly indicating that long-range information is more crucial for them.

For all datasets, we usually observe higher AUC for \posplus than its \pos variant suggesting the expressiveness power of center-common-neighbor pooling over simple center pooling. Of course, these slight AUC improvements come with slightly higher computational costs (see Table \ref{resource-cons}).

\vskip 1.5mm
\noindent \textbf{Results: OGB.} As shown in Table \ref{table:ogb-results}, our \posplus model can easily scale to graph datasets with millions of nodes and edges. Under the experimental setup in BUDDY, our \posplus still outperforms all the baselines (including BUDDY) for all Planetoid datasets, confirming the versatility of our model to different dataset splits and evaluation criteria. Our \posplus outperforms others on ogbl-collab, ogbl-citation2, and ogbl-vessel. For ogbl-ppa, we come in third place. For ogbl-ddi, \posplus performs significantly lower than SEAL and BUDDY; but we still perform better than the heuristic baselines. This performance on ogbl-ddi might be a limitation of our \posplus, possibly because the common-neighbor information is noisy in denser ogbl-ddi, with the performance degradation exaggerated by the lack of node feature information. However, this set of experiments confirms the competitive performance of \posplus on the OGB datasets with large-scale graphs under different types of efficacy measurements.

For OGB datasets, we also further compare the runtime of our \posplus with its main competitive counterparts SEAL. Table \ref{tab:time_OGB} reports an average preprocessing and inference time for an instance (i.e., target pair of link prediction). The inference times for \posplus are \grA{5--10} faster than those of SEAL with dynamic mode, where the maximum speedup is for PPA. \posplus also offers \grA{1.6--10} speedup over SEAL with the static mode.\footnote{In the static mode, SEAL generates and labels each subgraph a preprocessing whereas in the dynamic mode, each subgraph is generated on the fly during training.} The preprocessing time for SEAL static is slightly faster than that of \posplus with the \grA{1.2--1.6} speedup. However, we note that the SEAL static has a relatively low or no speedup over the SEAL dynamic; specifically, it doesn't even offer any speedup for PPA. This low or no speedup is consistent with our complexity analyses, demonstrating that both modes of SEAL have the same inference time complexity, which is dependent on the graph size. 
\begin{table}[tb]
\centering
\caption{Precomputed dataset sizes in Megabytes (MB) of SEAL vs. our \ssgrl models. \colorbox[rgb]{0.851,0.918,0.827}{Green} is the smallest and \colorbox[rgb]{0.957,0.8,0.8}{Red} is the largest dataset. Reduction \% = (\colorbox[rgb]{0.957,0.8,0.8}{Red} - \colorbox[rgb]{0.851,0.918,0.827}{Green}) / \colorbox[rgb]{0.957,0.8,0.8}{Red} $\times$ 100.}
\label{tab:space_req}
\scalebox{0.816}{
\begin{tabular}{lcccccc}
\toprule
&\multicolumn{3}{c}{\textbf{Cora}}&\multicolumn{3}{c}{\textbf{NS}}\\
Model&Training& Validation & Testing&Training& Validation & Testing \\
 \cmidrule[0.5pt](rl){1-1} \cmidrule[0.5pt](rl){2-4}\cmidrule[0.5pt](rl){5-7}
\pos & \best 786.44 & \best 23.05 & \best 46.18& \best 5.27 & \best 0.16 & \best 0.31 \\
\posplus & 920.93 & 26.81 & 53.83 & 8.83 & 0.27 & 0.53 \\
\sop & \best 786.44 & \best 23.05 & \best 46.18 & \best 5.27 & \best 0.16 & \best 0.31 \\
SEAL & \worst 19521.52 & \worst 620.09 & \worst 1172.68 & \worst 21.26 & \worst 0.64 & \worst 1.22 \\
 \cmidrule[0.5pt](rl){1-1} \cmidrule[0.5pt](rl){2-4}\cmidrule[0.5pt](rl){5-7}
Reduction \% & 95.97 & 96.28 & 96.06 & 75.21 & 75 & 74.59 \\
\bottomrule
&\multicolumn{3}{c}{\textbf{CiteSeer}}&\multicolumn{3}{c}{\textbf{Power}}\\ 
Model&Training& Validation & Testing&Training& Validation & Testing \\
 \cmidrule[0.5pt](rl){1-1} \cmidrule[0.5pt](rl){2-4}\cmidrule[0.5pt](rl){5-7}
\pos & \best 1750.53 & \best 51.34 & \best 102.91 & \best 12.66 & \best 0.37 & \best 0.75 \\
\posplus & 1996.66 & 57.90 & 117.38  & 13.27 & 0.39 & 0.79 \\
\sop & \best 1750.53 & \best 51.34 & \best 102.91& \best 12.66 & \best 0.37 & \best 0.75 \\
SEAL & \worst 18716.19 & \worst 500.03 & \worst 1083.71 & \worst 24.88 & \worst 0.72 & \worst 1.49 \\
\cmidrule[0.5pt](rl){1-1} \cmidrule[0.5pt](rl){2-4}\cmidrule[0.5pt](rl){5-7}
Reduction \% & 90.64 & 89.73 & 90.50 & 49.11 & 48.61 & 49.66 \\
\bottomrule
&\multicolumn{3}{c}{\textbf{PubMed}}&\multicolumn{3}{c}{\textbf{PB}}\\
Model&Training& Validation & Testing&Training& Validation & Testing \\
 \cmidrule[0.5pt](rl){1-1} \cmidrule[0.5pt](rl){2-4}\cmidrule[0.5pt](rl){5-7}
\pos & \best 2311.06 & \best 67.97 & \best 135.93 & \best 32.09 & \best 0.95 & \best 1.89 \\
\posplus & 2665.93 & 78.58 & 156.21& 137.19 & 4.04 & 7.99 \\
\sop & \best 2311.06 & \best 67.97 & \best 135.93 & \best 32.09 & \best 0.95 & \best 1.89 \\
SEAL & \worst 287324.77 & \worst 8500.97 & \worst 16670.80 & \worst 27814.36 & \worst 816.16 & \worst 1631.41 \\
\cmidrule[0.5pt](rl){1-1} \cmidrule[0.5pt](rl){2-4}\cmidrule[0.5pt](rl){5-7}
Reduction \% & 99.19 & 99.20 & 99.18 & 99.88 & 99.88 & 99.88 \\
\bottomrule
&\multicolumn{3}{c}{\textbf{Wisconsin}}&\multicolumn{3}{c}{\textbf{Yeast}}\\
Model&Training& Validation & Testing&Training& Validation & Testing \\
 \cmidrule[0.5pt](rl){1-1} \cmidrule[0.5pt](rl){2-4}\cmidrule[0.5pt](rl){5-7}
 \pos & \best 41.02 & \best 1.15 & \best 2.40 & \best 22.45 & \best 0.66 & \best 1.32 \\
 \posplus & 44.27 & 1.23 & 2.50& 80.65 & 2.37 & 4.75 \\
 \sop & \best 41.02 & \best 1.15 & \best 2.40 & \best 22.45 & \best 0.66 & \best 1.32 \\
 SEAL & \worst 349.59 & \worst 9.99 & \worst 20.04 & \worst 2321.44 & \worst 65.63 & \worst 135.84 \\
\cmidrule[0.5pt](rl){1-1} \cmidrule[0.5pt](rl){2-4}\cmidrule[0.5pt](rl){5-7}
Reduction \% & 88.26 & 88.48 & 88.02 & 99.03 & 98.99 & 99.02 \\
\bottomrule
\end{tabular}
}
\end{table}


\begin{table}[tb]
\centering
\addtolength{\extrarowheight}{\aboverulesep}
\caption{Results for \ssgrl as a scalability framework: ScaLed's subgraph sampling combined with \pos and \posplus.}
\label{table:ksup-scaled}
\scalebox{0.93}{
\begin{tabular}{p{3pt}|lcccc} 
\toprule
\textbf{}                 & \textbf{Model} & \textbf{Training} &  \textbf{Preproc.} & \textbf{Runtime} & \textbf{AUC}  \\ 
\hline
\multirow{4}{*}{\begin{sideways}
    Cora\end{sideways}}     & \pos           & 4.82{\scriptsize$\pm$0.25}               & 83.28{\scriptsize$\pm$3.16}      & 336.70{\scriptsize$\pm$2.47}    & 94.65{\scriptsize$\pm$0.67}  \\
                          & \pos  + ScaLed  & 4.73{\scriptsize$\pm$0.22}               & 60.32{\scriptsize$\pm$3.50}      & 308.73{\scriptsize$\pm$4.27}    & 94.35{\scriptsize$\pm$0.52}  \\
                          \cmidrule{2-6}
                          & \posplus           & 5.59{\scriptsize$\pm$0.31}               & 94.33{\scriptsize$\pm$7.36}      & 387.95{\scriptsize$\pm$7.81}    & 94.77{\scriptsize$\pm$0.65}  \\
                          & \posplus+ ScaLed  & 5.49{\scriptsize$\pm$0.28}             & 69.95{\scriptsize$\pm$5.12}      & 358.40{\scriptsize$\pm$7.08}    & 94.80{\scriptsize$\pm$0.58}  \\ 
\midrule
\multirow{4}{*}{\begin{sideways}CiteSeer \end{sideways}} & \pos             & 4.66{\scriptsize$\pm$0.20}              & 61.86{\scriptsize$\pm$0.64}      & 308.20{\scriptsize$\pm$2.86}    & 95.76{\scriptsize$\pm$0.59}  \\
                          & \pos + ScaLed  & 4.65{\scriptsize$\pm$0.21}               & 56.65{\scriptsize$\pm$2.59}      & 302.45{\scriptsize$\pm$3.95}    & 95.52{\scriptsize$\pm$0.65}  \\
                          \cmidrule{2-6}
                          & \posplus           & 4.72{\scriptsize$\pm$0.35}              & 79.14{\scriptsize$\pm$5.34}      & 330.29{\scriptsize$\pm$5.61}    & 95.60{\scriptsize$\pm$0.52}  \\
                          & \posplus+ ScaLed  & 4.66{\scriptsize$\pm$0.27}               & 65.56{\scriptsize$\pm$5.13}      & 313.38{\scriptsize$\pm$6.25}    & 95.60{\scriptsize$\pm$0.53}  \\
\bottomrule
\end{tabular}}
\end{table}

\vskip 1.5mm
\noindent \textbf{Dataset compression of \ssgrl vs SEAL.}
The preprocessing in SGRLs imposes additional disk space requirements to store preprocessed datasets including enclosing subgraphs for each target pair, and their augmented nodal features. Our study focuses on the disk space requirements of the final preprocessed datasets, as these represent the long-term storage needs for deploying SGRL models. Following up our theoretical analyses in Section \ref{sec:pos}, we intend to study how effective our S3GRL methods are in compressing the preprocessed datasets. Table \ref{tab:space_req} reports the disk space requirements of \ssgrl models vs SEAL and their effectiveness in disk space reduction. The \ssgrl has significantly reduced the space requirement, where the maximum is observed in PB in which \pos reduces the 28GB SEAL (preprocessed) training dataset to only 32MB dataset with the 99.88\% reduction.  For another example, \pos achieves a 99\% reduction in PubMed over SEAL: The SEAL training dataset requires 280GB of disk space in comparison to only about 2GB requirement in all S3GRL models.\footnote{Although we have not included an analysis of the intermediate storage requirements during preprocessing, we believe that the primary concern for practical applications is the disk space required to store the preprocessed datasets, as this impacts the feasibility of deploying and sharing models, especially on systems with limited storage capacity. Furthermore, since our methods process subgraphs sequentially without needing to hold large amounts of data in memory or on disk at once, the intermediate storage requirements are manageable and do not significantly affect the overall resource usage.}

\vskip 1.5mm
\noindent \textbf{\ssgrl as a scalability framework.}
To demonstrate the flexibility of \ssgrl as a scalability framework, we explore how easily it can host other scalable SGRLs. Specifically, we exchange the subgraph sampling strategy of \pos (and \posplus) with the random-walk induced subgraph sampling technique in \scaled. Through this process, we reduce our operator sizes and benefit from added regularization offered through stochasticity in random walks. We fixed its hyperparameters, the random walk length $h=3$, and the number of random walks $k=20$. Table \ref{table:ksup-scaled} shows the average computational time and AUC (over 10 runs).  The subgraph sampling of \scaled offers further speedup in \pos and \posplus in training, dataset preprocessing, and overall runtimes. For \posplus variants, we even witness AUC gains on Cora and no AUC losses on CiteSeer. The AUC gains of \posplus could be attributed to the regularization offered through sparser enclosing subgraphs. This demonstration suggests that \ssgrl can provide a foundation for hosting various SGRLs to further boost their scalability.

\section{Conclusions and Future Work}
Subgraph representation learning methods (SGRLs), albeit comprising state-of-the-art models for link prediction, suffer from large computational overheads. In this paper, we propose a novel SGRL framework \ssgrl, aimed at faster inference and training times while offering flexibility to emulate many other SGRLs. We achieve this speedup through easily precomputable subgraph-level diffusion operators in place of expensive message-passing schemes. \ssgrl supports multiple subgraph selection choices for the creation of the operators, allowing for a multi-scale view around the target links. Our experiments on multiple instances of our \ssgrl framework show significant computational speedup over existing SGRLs, while offering matching (or higher) link prediction efficacies. For future work, we intend to devise learnable subgraph selection and diffusion operators, that are catered to the training dataset and computational constraints.

\bibliographystyle{ACM-Reference-Format}
\bibliography{reference}

\end{document}